\documentclass[11pt]{article}

\usepackage[utf8]{inputenc}
\usepackage[T1]{fontenc}
\usepackage{amsmath,amssymb,amsfonts}
\usepackage{graphicx}
\usepackage{booktabs}
\usepackage[colorlinks=true,linkcolor=black,citecolor=blue,urlcolor=blue]{hyperref}
\usepackage{algorithm}
\usepackage{algorithmic}
\usepackage{xcolor}
\usepackage{geometry}
\usepackage{natbib}
\usepackage[font=footnotesize]{caption}
\usepackage{multirow}
\usepackage{listings}
\usepackage{pifont}
\usepackage{enumitem}
\usepackage{titlesec}
\usepackage{placeins}
\usepackage{float}

\title{RTMC: Step-Level Credit Assignment via Rollout Trees}

\author{
  Tao Wang \qquad Suhang Zheng \qquad Xiaoxiao Xu \\[8pt]
  Alibaba Group \\[6pt]
  {\small \texttt{\{shayue.wt, suhang.zhengsh, xiaoxiao.xuxx\}@alibaba-inc.com}}
}

\date{}

\begin{document}
\maketitle

\begin{abstract}
Multi-step agentic reinforcement learning benefits from fine-grained credit assignment, yet existing approaches offer limited options: critic-free methods like GRPO assign a uniform advantage to every action in a trajectory, while learned value networks introduce notable overhead and can be fragile under sparse rewards. We observe that group rollouts targeting the same problem often traverse overlapping intermediate states, implicitly forming a tree whose branches diverge at successive decision points. Building on this insight, we introduce Rollout-Tree Monte Carlo (RTMC) advantage estimation, which aggregates return statistics across rollouts sharing a common state to produce per-step Q-values and advantages---without any learned critic. A state-action signature system compresses raw interaction histories into compact, comparable representations, making cross-rollout state matching tractable. On SWE-bench Verified, RTMC improves pass@1 by 3.2 percentage points over GRPO.
\end{abstract}

\section{Introduction}
\label{sec:intro}

Large language models have achieved remarkable capabilities across a wide range of tasks~\citep{Qwen2025Qwen3, MiniMax2025MiniMax01, Kimi2025K15, DeepSeek2025R1}, yet most real-world problems cannot be solved by a single question-answer exchange. Such tasks require \textbf{agent systems} that pair an LLM with external tools---file editors, code executors, search utilities---and orchestrate multi-turn loops of reasoning and action~\citep{Yao2022React, Zhang2025Landscape}. 

Reinforcement learning has proven effective for improving such LLM-based agents~\citep{Ouyang2022Training, DeepSeek2025R1}. In single-turn settings, critic-free methods like GRPO~\citep{Shao2024DeepSeekMath} estimate advantages through group-level comparisons, eliminating the need for a learned value network. However, agentic RL~\citep{Zhang2025Landscape} is fundamentally different: the LLM acts as the decision-making core of an agent, and training resembles classic RL---with states, transitions, delayed rewards, and long horizons---where the consequence of a single tool call may not be apparent until many steps later.

\paragraph{Advantage estimation challenge in agentic RL.}
Applying RL to agentic tasks exposes a critical gap in credit assignment~\citep{Wang2025Practitioner, Zhang2025Agentrl}. Critic-free methods like GRPO assign a single advantage to an entire trajectory, unable to distinguish good steps from bad ones within a multi-turn episode. Training a critic via GAE~\citep{Schulman2016GAE} could help, but is expensive and struggles under sparse binary rewards. This has motivated a wave of recent work, including anchor-state grouping~\citep{Feng2025Group}, implicit process rewards~\citep{Liu2025Agentic}, tree-search-based credit assignment~\citep{Zong2026AT2PO}, retrospective critics~\citep{Zhang2025CriticSearch}, trajectory purification~\citep{Xu2026Cleaner}, and reflective retry~\citep{Shi2026R3L}. These methods either require additional learned components, introduce extra computation beyond the rollouts already collected, or do not provide true per-step advantage estimates.

\paragraph{Our key insight.}
Effective policy optimization fundamentally depends on the quality of advantage estimates---accurate per-step signals that distinguish beneficial actions from harmful ones~\citep{Sutton1999Policy, Kakade2002Approximately, Schulman2016GAE}. Continuing the critic-free paradigm, we cast agentic interaction as an MDP where each tool call is an action $a_t$ taken in state $s_t$, yielding a discounted return $G_t = \sum_{k=0}^{T-t} \gamma^k r_{t+k}$. When $N$ group rollouts attempt the same problem, they traverse different trajectories but may pass through similar states. This requires a way to recognize when two rollouts have reached the ``same'' state despite differing surface-level histories---a problem we address through state-action signatures (Section~\ref{sec:signatures}). Given such an abstraction, we can aggregate return statistics across rollouts that share a common state to estimate per-step advantages, without training any additional model.

\paragraph{Contributions.} We make the following contributions:
\newline
\begin{enumerate}
  \item \textbf{Rollout-tree Monte Carlo advantage estimation}: We propose aggregating return statistics from group rollouts over a tree structure to compute per-step Q-values and advantages, enabling fine-grained credit assignment without a critic network. To handle sparsely visited tree nodes, we introduce a prior-based value smoothing mechanism that ensures informative advantages throughout the tree (Section~\ref{sec:method}).

  \item \textbf{State-action signature system}: We design a state abstraction strategy that compresses raw agent interaction histories into compact, comparable signatures, making tree construction tractable. We instantiate this strategy for software engineering tasks via view bucketing, content hashing, and action classification (Section~\ref{sec:signatures}).

  \item \textbf{Empirical validation}: On SWE-bench Verified, a progressive ablation from the untrained baseline through GRPO, step-level rewards, and our tree-based estimator demonstrates additive gains at each stage, with Rollout-Tree MC achieving 52.2\% pass@1---a 3.2 percentage point improvement over GRPO (Section~\ref{sec:experiments}).
\end{enumerate}

\section{Related Work}
\label{sec:related}

LLM training has progressed from supervised fine-tuning~\citep{Ouyang2022Training} through RLHF~\citep{Christiano2017Deep, Ouyang2022Training, Schulman2017PPO} to reinforcement learning with verifiable rewards (RLVR), where rule-based signals replace learned reward models~\citep{DeepSeek2025R1, Shao2024DeepSeekMath}. The natural next frontier is \textbf{agentic RL}~\citep{Zhang2025Landscape}: using RL to improve LLM policies within multi-turn agent systems that pair an LLM with external tools and an environment~\citep{Wang2023Survey, Yao2022React}.

\paragraph{Single-turn RLVR.}
In RLVR, a policy $\pi_\theta$ generates a response $y$ for prompt $x$ and receives a verifiable reward $R(x, y)$~\citep{DeepSeek2025R1}. GRPO~\citep{Shao2024DeepSeekMath} samples $K$ responses per prompt and computes
\begin{equation}
  A_i = \frac{R(x, y_i) - \mu}{\sigma}, \quad \mu = \frac{1}{K}\sum_{j=1}^{K} R(x, y_j), \quad \sigma = \mathrm{std}\bigl(\{R(x, y_j)\}_{j=1}^{K}\bigr).
  \label{eq:grpo}
\end{equation}
This advantage is shared across all tokens within response $y_i$. The approach is effective for single-turn tasks but provides no mechanism to distinguish good steps from bad ones within a multi-step episode.

\paragraph{Multi-turn agentic RL as an MDP.}
Agentic tasks are fundamentally sequential: an agent interacts with an environment over multiple turns, and the consequence of a single action may not manifest until many steps later~\citep{Zhang2025Landscape, Wang2025Practitioner}. We formalize this as a Markov Decision Process $(\mathcal{S}, \mathcal{A}, P, R, \gamma)$:
\begin{itemize}
  \item \textbf{State} $s_t \in \mathcal{S}$: the agent's problem-solving progress at step $t$.
  \item \textbf{Action} $a_t \in \mathcal{A}$: an LLM-generated tool call (e.g., file editor, code executor, shell command).
  \item \textbf{Transition} $P(s_{t+1} \mid s_t, a_t)$: the environment executes $a_t$ and returns the next state.
  \item \textbf{Reward} $R \in \{0, 1\}$: a sparse terminal signal for task success.
  \item \textbf{Discount factor} $\gamma \in (0, 1]$: trades off immediate and future rewards~\citep{Sutton2018Reinforcement}.
\end{itemize}

The discounted return from step $t$ is $G_t = \sum_{k=0}^{T-t} \gamma^k r_{t+k}$. Following the standard RL framework~\citep{Sutton2018Reinforcement, Sutton1999Policy}, we define the action-value function $Q$, the state-value function $V$, and the advantage function $A$:
\begin{equation}
  Q(s, a) = \mathbb{E}[G_t \mid s_t = s, a_t = a], \quad V(s) = \mathbb{E}_{a \sim \pi}[Q(s, a)], \quad A(s, a) = Q(s, a) - V(s).
  \label{eq:mdp-qva}
\end{equation}
Accurate advantage estimation is central to policy gradient methods~\citep{Schulman2016GAE, Kakade2002Approximately}. Unlike single-turn RLVR, agentic trajectories consist of heterogeneous steps that contribute unevenly to the outcome; distributing a terminal reward uniformly is problematic~\citep{Feng2025Group, Liu2025Agentic}.

\paragraph{Credit assignment and advantage estimation.}
Recent work addresses the credit assignment gap in agentic RL through various mechanisms.
GiGPO~\citep{Feng2025Group} groups actions from matching anchor states across rollouts, but requires exact raw-state matching, which rarely occurs in complex environments.
iStar~\citep{Liu2025Agentic} trains an implicit process reward model that infers step-level rewards from trajectory preferences, introducing a separate learned component into the training pipeline.
CriticSearch~\citep{Zhang2025CriticSearch} employs a retrospective critic for dense feedback, but also requires training an additional model.
AT$^2$PO~\citep{Zong2026AT2PO} propagates terminal rewards through a tree, but requires additional rollouts beyond those already collected.
CLEANER~\citep{Xu2026Cleaner} purifies trajectories at the data level but retains trajectory-level GRPO advantages.
R$^3$L~\citep{Shi2026R3L} identifies divergence points via reflect-then-retry but requires extra inference and does not provide per-step advantages.
Closer to our work are methods that use tree or branching structures to estimate per-step values.
VinePPO~\citep{Kazemnejad2025VinePPO} launches branch rollouts from each intermediate state to obtain unbiased MC value estimates, but requires extra inference proportional to sequence length.
TEMPO~\citep{Tran2025TEMPO} builds a prefix tree over group responses and applies branch-gated TD corrections, remaining critic-free with no extra rollouts. It treats raw token prefixes as state identity, however, which assumes that equivalent problem-solving progress produces identical token histories---an assumption that rarely holds in agentic settings.
Our method requires no additional inference or learned components: by decoupling state identity from raw context through an explicit signature system, we enable cross-rollout tree construction and unbiased per-step advantage estimation precisely where prefix-based approaches fail.

\section{Rollout-Tree Monte Carlo Advantage Estimation}
\label{sec:method}

Training a critic for LLM agents is difficult due to combinatorially large state spaces and sparse binary rewards~\citep{Wang2025Practitioner}. GRPO~\citep{Shao2024DeepSeekMath} avoids a critic by using group-level returns, but assigns all steps the same advantage. Our goal is to preserve GRPO's critic-free efficiency while recovering step-level discrimination.

The key observation is that $N$ group rollouts for the same problem already constitute natural counterfactual comparisons: rollouts sharing a common state prefix but diverging at a decision point reveal the causal effect of each branching action. We organize these rollouts into a tree via \emph{state-action signatures} (Section~\ref{sec:signatures}) and apply first-visit Monte Carlo estimation~\citep{Sutton2018Reinforcement} to obtain unbiased per-step advantages without any additional model or computation.

\subsection{From Rollouts to Trees}
\label{sec:method-tree}

A trajectory $\tau$ is an alternating sequence of states, actions, and rewards:
\begin{equation}
  \tau = (s_0, a_0, r_0, \; s_1, a_1, r_1, \; \ldots, \; s_T),
  \label{eq:traj}
\end{equation}
where $s_t$ is the state, $a_t$ the action, and $r_t$ the reward~\citep{Sutton2018Reinforcement}. In the LLM agentic setting, each action $a_t$ is a variable-length token span (reasoning trace + tool call + arguments), and the state $s_t$ is the accumulated textual context. The rollout unfolds as:
\begin{equation}
  \tau = \bigl(\underbrace{u_1, \ldots, u_{p_0}}_{s_0}, \;\underbrace{w_{b_0}, \ldots, w_{e_0}}_{a_0}, \;r_0, \;\;\underbrace{u_1, \ldots, u_{p_1}}_{s_1}, \;\underbrace{w_{b_1}, \ldots, w_{e_1}}_{a_1}, \;r_1, \;\;\ldots\bigr),
  \label{eq:traj-token}
\end{equation}
Only the action segments $a_t$ participate in the policy gradient; state context tokens are masked. Our advantage estimation assigns a single value to each $(s_t, a_t)$ pair, broadcast to all tokens in the action segment.

To organize $N$ rollouts into a tree, we need cross-rollout comparability of states and actions. We achieve this through \emph{state-action signatures} (Section~\ref{sec:signatures}), which abstract the cumulative interaction history into compact identifiers. Because signatures encode cumulative history, each appears at most once per rollout, satisfying the first-visit Monte Carlo condition.

\begin{figure}[t]
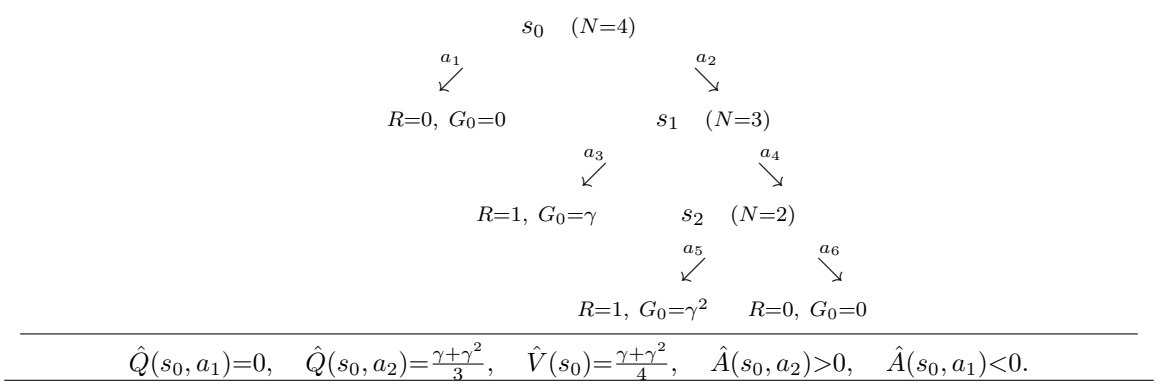

\centering
\begin{tabular}{c}
\hline\\[-6pt]
\begin{minipage}{0.90\textwidth}
\small\centering
$s_0$ \; {\scriptsize($N{=}4$)} \\[4pt]
$\overset{\scriptstyle a_1}{\swarrow}$ \hspace{80pt} $\overset{\scriptstyle a_2}{\searrow}$ \\[3pt]
{\scriptsize$R{=}0,\; G_0{=}0$} \hspace{50pt} $s_1$ \; {\scriptsize($N{=}3$)} \\[5pt]
\hspace{75pt} $\overset{\scriptstyle a_3}{\swarrow}$ \hspace{50pt} $\overset{\scriptstyle a_4}{\searrow}$ \\[3pt]
\hspace{40pt} {\scriptsize$R{=}1,\; G_0{=}\gamma$} \hspace{25pt} $s_2$ \; {\scriptsize($N{=}2$)} \\[5pt]
\hspace{135pt} $\overset{\scriptstyle a_5}{\swarrow}$ \hspace{35pt} $\overset{\scriptstyle a_6}{\searrow}$ \\[3pt]
\hspace{105pt} {\scriptsize$R{=}1,\; G_0{=}\gamma^2$} \hspace{8pt} {\scriptsize$R{=}0,\; G_0{=}0$} \\[5pt]
\hrule\vspace{3pt}
$\hat{Q}(s_0, a_1) {=} 0$, \;\;
$\hat{Q}(s_0, a_2) {=} \tfrac{\gamma + \gamma^2}{3}$, \;\;
$\hat{V}(s_0) {=} \tfrac{\gamma + \gamma^2}{4}$, \;\;
$\hat{A}(s_0, a_2) {>} 0$, \;\;
$\hat{A}(s_0, a_1) {<} 0$.
\end{minipage} \\[4pt]
\hline
\end{tabular}
\caption{A rollout tree from four rollouts. Rollout~1 terminates at depth~1 via $a_1$ (failure), while rollouts~2--4 share action $a_2$ and diverge at deeper states $s_1, s_2$. Returns $G_0$ are discounted from terminal rewards back to $s_0$, yielding per-action advantages.}
\label{fig:rollout-tree}
\end{figure}

Given $N$ rollouts for the same problem, we construct the rollout tree in a single pass. For each step $t$ in each rollout $\tau^{(i)}$, we compute the discounted future return:
\begin{equation}
  G_t^{(i)} = \sum_{k=0}^{T_i - t} \gamma^k \, r_{t+k}^{(i)},
  \label{eq:return}
\end{equation}
and update a statistics table $\mathcal{T}$ indexed by state-action pairs:
\begin{equation}
  \mathcal{T}(s, a) = \Bigl\{ N(s, a), \;\; \textstyle\sum_{\text{visits}} G_t^{(i)} \Bigr\},
  \label{eq:stats}
\end{equation}
where $N(s, a)$ counts the number of rollouts that visit state $s$ and take action $a$, and the sum accumulates their returns. This is precisely the bookkeeping required for first-visit Monte Carlo estimation~\citep{Sutton2018Reinforcement}. Figure~\ref{fig:rollout-tree} illustrates this process on a concrete example.

\subsection{Monte Carlo Advantage Computation}
\label{sec:method-advantage}

We use hat notation ($\hat{Q}$, $\hat{V}$, $\hat{A}$) for Monte Carlo estimates, distinguishing them from the true values in Eq.~\eqref{eq:mdp-qva}.

\paragraph{Action-value estimate.} The first-visit MC estimator~\citep{Sutton2018Reinforcement} averages returns across rollouts visiting $(s, a)$:
\begin{equation}
  \hat{Q}(s, a) = \frac{1}{N(s,a)} \sum_{i : (s_t^{(i)}, a_t^{(i)}) = (s, a)} G_t^{(i)}.
  \label{eq:qhat}
\end{equation}
Each $G_t^{(i)}$ is an independent, unbiased sample of $Q^\pi(s, a)$ under the current policy $\pi$, so $\hat{Q}(s, a)$ is an unbiased estimator with variance $\mathrm{Var}[\hat{Q}] = \sigma^2_{Q}(s,a) / N(s,a)$.

\paragraph{State-value estimate.} The visit-weighted average of action values equals the sample mean of all returns from $s$:
\begin{equation}
  \hat{V}(s) = \frac{\sum_{a \in \mathcal{A}_s} N(s,a) \cdot \hat{Q}(s,a)}{\sum_{a \in \mathcal{A}_s} N(s,a)} = \frac{1}{N(s)} \sum_{i : s_t^{(i)} = s} G_t^{(i)},
  \label{eq:vhat}
\end{equation}
where $N(s) = \sum_{a} N(s, a)$. The visit-weighted form recovers $V^\pi(s)$ since visit frequencies reflect the policy's action distribution.

\paragraph{Advantage estimate.}
\label{sec:method-token}
The per-step advantage is the difference between the action value and the state value:
\begin{equation}
  \hat{A}(s, a) = \hat{Q}(s, a) - \hat{V}(s).
  \label{eq:ahat}
\end{equation}
Since each action $a_t$ spans a contiguous token segment $[b_t, e_t]$ (Eq.~\eqref{eq:traj-token}), the step-level advantage is broadcast to all tokens in that segment. Following the PPO clipped surrogate objective~\citep{Schulman2017PPO}:
\begin{equation}
  \mathcal{L}(\theta) = -\mathbb{E}\Bigl[\sum_{t} \sum_{j=b_t}^{e_t} \min\!\bigl(r_j(\theta)\,\hat{A}(s_t, a_t),\;\mathrm{clip}\bigl(r_j(\theta),\,1{-}\epsilon,\,1{+}\epsilon\bigr)\hat{A}(s_t, a_t)\bigr)\Bigr],
  \label{eq:loss}
\end{equation}
where $r_j(\theta) = \dfrac{\pi_\theta(w_j \mid \text{context}_{<j})}{\pi_{\theta_{\text{old}}}(w_j \mid \text{context}_{<j})}$ is the token-level probability ratio and $\epsilon$ is the clipping threshold.

\subsection{Procedure and Properties}
\label{sec:method-algorithm}

Algorithm~\ref{alg:mcts} summarizes the procedure. Two linear passes over the rollout data suffice: the first populates the statistics table $\mathcal{T}$, the second reads off per-step advantages and broadcasts them to the corresponding token segments. The overall complexity is $O(N \cdot L)$ in scalar operations, where $N$ is the number of rollouts and $L$ the average episode length, with no gradient computation involved.

\begin{algorithm}[t]
\caption{Rollout-Tree Monte Carlo Advantage Estimation}
\label{alg:mcts}
\small
\begin{algorithmic}[1]
\REQUIRE Batch of rollouts grouped by \texttt{problem\_id}
\ENSURE Token-level advantages
\FOR{each \texttt{problem\_group} in batch}
  \STATE $\mathcal{T} \leftarrow \text{new StatisticsTable}$
  \STATE \textbf{// Phase 1: Build rollout tree}
  \FOR{each rollout $\tau^{(i)}$ in \texttt{problem\_group}}
    \FOR{each step $t = 0, 1, \ldots, T_i$}
      \STATE $s \leftarrow \textsc{StateSignature}(\text{history}_{<t})$
      \STATE $a \leftarrow \textsc{ActionSignature}(\text{step}_{t})$
      \STATE $G_t \leftarrow \sum_{k=0}^{T_i - t} \gamma^k \, r_{t+k}$
      \STATE $\mathcal{T}(s, a).\text{count} \mathrel{+}= 1$; \quad $\mathcal{T}(s, a).\text{sum} \mathrel{+}= G_t$
    \ENDFOR
  \ENDFOR
  \STATE \textbf{// Phase 2: Compute per-step advantages}
  \FOR{each rollout $\tau^{(i)}$ in \texttt{problem\_group}}
    \FOR{each step $t = 0, 1, \ldots, T_i$}
      \STATE $\hat{Q} \leftarrow \mathcal{T}(s, a).\text{sum}\; / \;\mathcal{T}(s, a).\text{count}$ \hfill $\triangleright$ Eq.~\eqref{eq:qhat}
      \STATE $\hat{V} \leftarrow \sum_{a'} \mathcal{T}(s, a').\text{sum}\; / \;\sum_{a'} \mathcal{T}(s, a').\text{count}$ \hfill $\triangleright$ Eq.~\eqref{eq:vhat}
      \STATE $\hat{A}_t \leftarrow \hat{Q} - \hat{V}$ \hfill $\triangleright$ Eq.~\eqref{eq:ahat}
    \ENDFOR
    \STATE Broadcast $\hat{A}_t$ to tokens $[b_t, e_t]$ for each step $t$
  \ENDFOR
  \STATE Divide advantages by group-level std (optional per-group normalization)
\ENDFOR
\RETURN all token-level advantages
\end{algorithmic}
\end{algorithm}

\paragraph{Per-group normalization.}
Because different problems produce advantages at different scales (e.g., easy problems have near-uniform returns while hard problems exhibit high variance), we optionally normalize advantages within each problem group by dividing by the group-level standard deviation computed over step-level advantages. This prevents high-variance groups from dominating the gradient while preserving the sign and relative ordering of advantages within each group.

\paragraph{Properties.}
\label{sec:method-theory}
By the first-visit MC guarantee~\citep{Sutton2018Reinforcement}, $\hat{Q}$, $\hat{V}$, and $\hat{A}$ are all unbiased. Variance scales as $\sigma^2_Q / N(s,a)$: root-level states have low variance while leaf states have high variance, motivating prior smoothing (Section~\ref{sec:corrections}). The method generalizes GRPO along the axis of state abstraction granularity: at the coarsest extreme (all rollouts share a single root state) the tree collapses to trajectory-level GRPO, and it is critic-free unlike GAE.

\subsection{Corrections to the Advantage Estimator}
\label{sec:corrections}

Two practical issues arise when applying this estimator to agentic tasks. First, states visited by only a single rollout yield $\hat{A} = 0$ regardless of action quality, because $\hat{Q} = \hat{V}$ when only one action is observed. Second, in coding tasks usually requiring validation, such as SWE, the discount factor $\gamma$ penalizes multi-step verification actions. We address each below.

\paragraph{Prior-based value smoothing.}
\label{sec:prior}
In practice, a large fraction of states are visited by only a single rollout, an inherent limitation of tree-based Monte Carlo with finite $N$. We regularize by blending the local estimate with a global prior:
\begin{equation}
  \hat{V}'(s) = \frac{n(s) \cdot \hat{V}(s) + n_{\text{prior}} \cdot V_{\text{prior}}}{n(s) + n_{\text{prior}}}
  \label{eq:prior}
\end{equation}
where $V_{\text{prior}}$ is the group success rate and $n_{\text{prior}} = 2$. This Bayesian shrinkage~\citep{JamesStein1961} pulls low-sample estimates toward the group mean; at $n{=}1$ the prior contributes 67\%, vanishing as visits grow.

\paragraph{Task-specific reward shaping.}
\label{sec:vrs}
An agent finishing immediately gets $\hat{Q} = 1.0$, while one running tests first gets $\hat{Q} \approx 0.98$, receiving negative advantage despite best practice. We add a small reward to validation actions after code modifications:
\begin{equation}
  r'_t = r_t + \beta \cdot \mathbb{1}[\text{has\_modifications}(s_t)] \cdot \mathbb{1}[\text{is\_validation}(a_t)]
  \label{eq:vrs}
\end{equation}
where $\beta = 0.05$. This is potential-based reward shaping~\citep{Ng1999Reward}, preserving the optimal policy. Other agentic domains may require analogous domain-specific corrections or none at all.

\section{State-Action Signature Design}
\label{sec:signatures}

We instantiate the framework on SWE tasks~\citep{Jimenez2024SWEbench}, where an agent following the SWE-agent architecture~\citep{Yang2024Swe} interacts with code repositories via R2E-Gym tools~\citep{Jain2025R2EGym}: file editor, search, bash executor, think, and finish. Action types are finite but arguments are unbounded.

Two properties make na\"ive representations inadequate: (1)~\emph{combinatorial state space}---raw interaction histories almost never match across rollouts even when problem-solving progress is equivalent; (2)~\emph{functional overlap}---bash commands subsume dedicated tools (e.g., \texttt{cat} $\equiv$ \texttt{file\_editor view}), fragmenting the tree if recorded literally. We therefore design \emph{signatures} that map this space into compact, comparable representations. Signature granularity governs tree quality: too coarse mixes heterogeneous returns, too fine reduces to trajectory-level estimation.

\subsection{Action Signature}

We classify tool invocations by \emph{effect on progress} rather than tool name (Table~\ref{tab:action-taxonomy}). Equivalent tools map to the same category.

\begin{table}[t]
\centering\small
\caption{Action categories classified by effect on problem-solving progress. Dedicated tools and bash commands with equivalent effects map to the same category.}
\label{tab:action-taxonomy}
\begin{tabular}{@{}lll@{}}
\toprule
Category & Description & Example \\
\midrule
\texttt{view} & Inspect file content (full or range-based) & \texttt{file\_editor view}, \texttt{cat}, \texttt{head} \\
\texttt{search} & Locate files or code patterns & \texttt{search}, \texttt{grep}, \texttt{find}, \texttt{ls} \\
\texttt{modify} & Edit existing code (replace or insert) & \texttt{str\_replace}, \texttt{insert} \\
\texttt{create} & Create a new file & \texttt{file\_editor create} \\
\texttt{execute} & Run a script for debugging or reproduction & \texttt{python reproduce.py} \\
\texttt{test} & Execute test suites to validate changes & \texttt{pytest}, \texttt{python -m unittest} \\
\texttt{install} & Manage dependencies & \texttt{pip install} \\
\texttt{fileop} & File-system operations (copy, move, delete) & \texttt{cp}, \texttt{mv}, \texttt{mkdir} \\
\texttt{think} & Internal reasoning without environment interaction & \texttt{think} \\
\texttt{finish} & Submit the solution and terminate the episode & \texttt{finish} \\
\bottomrule
\end{tabular}
\end{table}

Each action is encoded as a structured string that captures not only the category but also the \emph{scope} of the operation.
The general format is \texttt{category:scope@target[:result]}, where \texttt{category} is drawn from Table~\ref{tab:action-taxonomy}, \texttt{scope} encodes operation-specific detail, \texttt{target} is the primary file, and \texttt{result} records success or error when applicable.
For \texttt{view} actions, the scope distinguishes full-file reads (\texttt{view:full@core.py}) from range-based reads with 100-line bucket identifiers (\texttt{view:partial[1\text{-}2]@core.py} for lines 100--299).
For \texttt{modify} actions, the scope includes a 4-character content hash of the changed text (\texttt{modify:replace:a3f2@core.py}), so that different edits to the same file produce distinct signatures.
For \texttt{test} and \texttt{execute} actions, the result suffix distinguishes successful runs from failures (\texttt{test@test\_main.py:ok} vs.\ \texttt{test@test\_main.py:error}).
\texttt{think} and \texttt{finish} actions carry no file target and are encoded as bare categories.

\subsection{State Signature}

A state signature captures the \emph{cumulative effect} of all preceding actions by step~$t$: which files have been inspected, what edits applied, and what execution outcomes observed. It must abstract surface differences (e.g., file access order) while preserving progress distinctions. We encode this as an ordered, per-file record:
\begin{equation}
  s = \texttt{file}_1\texttt{:OPS}_1 \;\mid\; \texttt{file}_2\texttt{:OPS}_2 \;\mid\; \ldots \;\mid\; \texttt{(FLAGS)}
\end{equation}

Each file contributes \texttt{file:OPS}, where \texttt{OPS} is the \emph{set} (not sequence) of operations performed (Table~\ref{tab:state-ops}), ensuring order-invariance. Non-file actions are recorded in \texttt{FLAGS} (think count, test pass/fail counts). Files are sorted alphabetically for determinism.

\begin{table}[b]
\centering\small
\caption{State signature operation encodings.}
\label{tab:state-ops}
\begin{tabular}{@{}llll@{}}
\toprule
Operation & Encoding & Example & Notes \\
\midrule
View (full file) & \texttt{Vf} & \texttt{core.py:Vf} & \\
View (partial) & \texttt{V[b]} & \texttt{core.py:V[2]} & Lines 200--299 \\
Modify & \texttt{M:xxxx} & \texttt{core.py:M:a3f2} & 4-char content hash \\
Insert & \texttt{I:xxxx} & \texttt{core.py:I:b7c1} & 4-char content hash \\
Search & \texttt{S} & \texttt{core.py:S} & \\
Create & \texttt{C} & \texttt{test.py:C} & \\
\bottomrule
\end{tabular}
\end{table}

\paragraph{View bucketing.}
We partition files into 100-line buckets, recording only bucket indices (\texttt{V[1-2]} for lines 100--299). This bounds distinct view states to $\lceil L / 100 \rceil$ per file while retaining coarse location sensitivity.

\paragraph{Content hashing.}
Each edit is represented by a 4-character MD5 hash of the concatenated old and new text, yielding $16^4 = 65{,}536$ possible hashes per file---sufficient to avoid collisions while keeping signatures compact.

\section{Experimental Analysis}
\label{sec:experiments}

\subsection{Experimental Setup}
\label{sec:exp-setup}

We train Qwen3-Coder-30B-A3B-Instruct~\citep{Qwen2025Qwen3} with the ROLL framework~\citep{Wang2025ROLL} on 32 GPUs, and evaluate on SWE-bench Verified~\citep{Jimenez2024SWEbench} (500 GitHub issues). The agent follows SWE-agent~\citep{Yang2024Swe} in the ROCK sandbox~\citep{Wang2025ROCK} with R2E-Gym tools~\citep{Jain2025R2EGym}. Training data is an 8.1K-instance R2E-Gym subset~\citep{Jain2025R2EGym} with no SWE-bench overlap. The metric is \emph{pass@1} determined by majority vote over 3 runs.

We compare four configurations in a progressive ablation:
\begin{itemize}
  \item \textbf{Baseline}: pretrained model without RL training;
  \item \textbf{GRPO}: trajectory-level advantage from terminal reward only~\citep{Shao2024DeepSeekMath};
  \item \textbf{GRPO + Step Reward}: GRPO augmented with per-step rewards ($+0.005$ valid, $-0.005$ invalid, $+0.05$ validation) accumulated via $\gamma$-discounted returns;
  \item \textbf{Rollout-Tree MC (Ours)}: tree-based per-step advantages ($\gamma{=}0.99$, prior smoothing $n_{\text{prior}}{=}2$, validation bonus $\beta{=}0.05$), using the same step rewards.
\end{itemize}
All RL configurations share: learning rate $1{\times}10^{-7}$ with a linear constant scheduler and 20-step warmup, 8 problems $\times$ 8 rollouts = 64 trajectories per step, filtering of uniform-outcome groups, and off-policy minibatch updates with loss scaling.

\subsection{Main Results}
\label{sec:main-results}

As shown in Table~\ref{tab:main-results}, each component in our progressive ablation yields additive gains: episode-level GRPO improves over the baseline model, adding step-level rewards provides further improvement, and Rollout-Tree MC achieves the best performance through tree-based per-step advantage estimation.

The absolute gains are modest---each stage contributes between 1.4 and 2.2 percentage points---but the strictly monotonic ordering across all four configurations is notable: every refinement in credit assignment granularity (episode-level $\to$ step-level $\to$ state-level) translates into a measurable improvement.

\begin{table}[H]
\centering\small
\caption{Pass@1 on SWE-bench Verified (500 problems), determined by majority vote over 3 evaluation runs. Best checkpoint selected for each method. $\Delta$ denotes improvement over the baseline.}
\label{tab:main-results}
\begin{tabular}{@{}lccc@{}}
\toprule
Method & Resolved & Pass@1 & $\Delta$ vs Baseline \\
\midrule
Baseline (no training) & 234 / 500 & 46.8\% & --- \\
GRPO & 245 / 500 & 49.0\% & +2.2 pp \\
GRPO + Step Reward & 252 / 500 & 50.4\% & +3.6 pp \\
Rollout-Tree MC & \textbf{261 / 500} & \textbf{52.2\%} & \textbf{+5.4 pp} \\
\bottomrule
\end{tabular}
\end{table}

\subsection{Learning Curves}
\label{sec:training-dynamics}

\begin{figure}[H]
\centering
\includegraphics[width=0.85\linewidth]{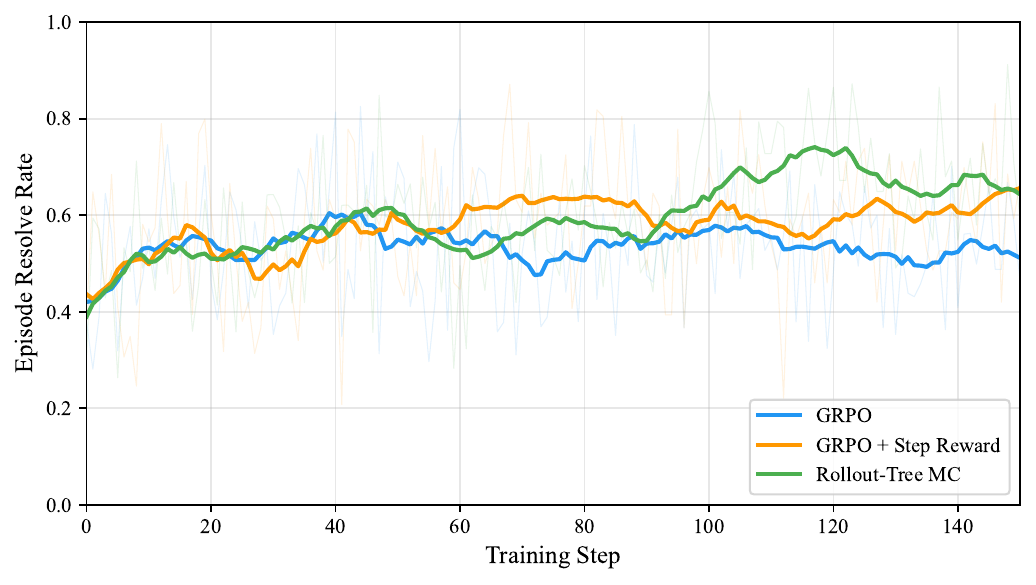}
\caption{Training episode resolve rate on the R2E dataset. All methods start from the same pretrained checkpoint and share the GRPO optimization framework; they differ only in how per-token advantages are computed. The progressive separation confirms that finer-grained credit assignment yields more effective policy gradients.}
\label{fig:training-curves}
\end{figure}

Figure~\ref{fig:training-curves} tracks the episode resolve rate on the R2E training set across 150 optimization steps. Note that this metric is computed over \emph{all} rollout groups, including those with uniform outcomes (all success or all failure) that are excluded from gradient updates (Section~\ref{sec:exp-setup}), and therefore reflects the policy's overall problem-solving ability rather than the training signal quality. The three methods separate progressively: episode-level GRPO learns slowly under high-variance trajectory signals, step-level reward decomposition narrows the credit assignment gap and accelerates early improvement, and tree-based advantage estimation widens the margin further by enabling state-level action comparison. This ordering---consistent throughout training---mirrors the ablation hierarchy in Table~\ref{tab:main-results} and supports the hypothesis that finer-grained credit assignment translates into more effective policy gradients.

\subsection{Ablation Study}
\label{sec:ablation}

We isolate the contribution of prior-based value smoothing by comparing the full Rollout-Tree MC configuration against a variant trained without the prior ($n_{\text{prior}}{=}0$), with all other hyperparameters held constant.

\begin{table}[b]
\centering
\caption{Ablation of prior-based value smoothing. Removing the prior ($n_{\text{prior}}{=}0$) reduces pass@1 by 2.5 percentage points.}
\label{tab:ablation-prior}
\begin{tabular}{@{}lc@{}}
\toprule
Configuration & Pass@1 \\
\midrule
Rollout-Tree MC w/o prior ($n_{\text{prior}}{=}0$) & 49.7\% \\
Rollout-Tree MC full ($n_{\text{prior}}{=}2$) & \textbf{52.2\%} \\
\midrule
Improvement from prior smoothing & +2.5 pp \\
\bottomrule
\end{tabular}
\end{table}

Table~\ref{tab:ablation-prior} shows that removing prior smoothing reduces pass@1 from 52.2\% to 49.7\%, a 2.5 pp drop that falls below GRPO + Step Reward (50.4\%). This confirms that the prior is essential: as noted in Section~\ref{sec:corrections}, a large fraction of states are visited only once and produce zero advantage without smoothing. The prior converts these dead nodes into informative training signals by anchoring their value estimates to the group-level success rate.

\section{Discussion}
\label{sec:discussion}

\subsection{Advantages of the Approach}

\textbf{Group-informed credit assignment.}
Unlike trajectory-level methods (GRPO) that assign identical advantages to all actions within a trajectory, and step-reward methods where credit remains a function of a single trajectory, Rollout-Tree MC determines each action's advantage by comparing its return against outcomes of \emph{all} rollouts sharing the same state. This enables within-trajectory differentiation---penalizing suboptimal actions in successful trajectories and rewarding good actions in failed ones---while grounding advantages in relative action quality at each decision point rather than absolute returns subject to problem-difficulty confounds.

\textbf{Critic-free and lightweight.} The method computes advantages purely from rollout return statistics, eliminating the need to train and maintain a separate value network. Tree construction and advantage computation are linear in the number of rollouts and episode length, requiring no gradient computation or additional inference beyond the rollouts already collected for policy optimization.

\subsection{Limitations and Future Work}

While Rollout-Tree MC consistently outperforms both GRPO baselines, the absolute improvement remains modest (5.4 pp over the untrained baseline, 1.8 pp over GRPO + Step Reward). In essence, our approach trades a learned critic for engineered state abstractions: it eliminates the cost of training and maintaining a value network, but shifts the burden to domain-specific signature design. We attribute the current performance ceiling primarily to the hand-crafted nature of this signature system.

\textbf{Signature coverage and granularity.}
Our signature design relies on manually defined rules---action classification, view bucketing, content hashing---that inevitably leave blind spots. Actions whose effects are difficult to capture through surface-level heuristics (e.g., subtle differences in reasoning depth) may be mapped to identical signatures, collapsing distinct states into one tree node. Conversely, overly fine distinctions fragment the tree and reduce visit counts. With $N{=}8$ rollouts per problem, the majority of states are visited only once, limiting where true cross-rollout action comparison can occur.

\textbf{State isolation.}
Because most tree nodes receive only a single visit, the raw MC estimate yields zero advantage there; the prior compensates but cannot substitute for genuine multi-action comparisons. Increasing the proportion of shared states---through more rollouts, coarser signatures, or both---is the most direct path to stronger credit assignment. The validation reward shaping we use is specific to SWE tasks where verification steps are penalized by $\gamma$-discounting; other domains may need different corrections or none at all.

\textbf{Toward learned signatures.}
One promising direction is replacing hand-crafted signatures with learned embeddings from a pre-trained encoder, clustering semantically similar states in a continuous space. This could capture similarities that rule-based signatures miss---for instance, recognizing that different debugging strategies have reached equivalent progress---but reintroduces dependency on an external model. Crucially, the tree-based estimation framework itself is agnostic to how signatures are produced: as signature quality improves or rollout budgets grow, the proportion of shared tree nodes increases and the method's advantage over flat baselines widens, suggesting that the current gains represent a lower bound on the approach's potential.

\textbf{Performance redistribution.} We observe that RL training does not uniformly improve performance: the trained model solves new problems that the base checkpoint fails on, but also regresses on a subset of previously solved instances. Understanding and mitigating this regression---whether through better credit assignment, curriculum design, or checkpoint selection---remains an open question.

\textbf{Single benchmark.} We evaluate on SWE-bench Verified with a single base model. Experiments on additional benchmarks, task domains, and model families are needed to establish broader generality.

\section{Conclusion}
\label{sec:conclusion}

We presented Rollout-Tree MC, a critic-free advantage estimator that organizes group rollouts into a shared tree to enable step-level credit assignment for agentic RL. Experiments on SWE-bench Verified show consistent gains over trajectory-level and step-level baselines, with each stage of credit assignment refinement contributing to the overall improvement. The approach adds negligible overhead and integrates directly into existing GRPO-based pipelines. The main limitation---hand-crafted state-action signatures---suggests that combining rule-based abstraction with learned representations is a promising direction for scaling tree-based advantage estimation to broader agentic domains.

\section*{Acknowledgments}

Appreciation is extended to Qifang Zhao and Liang Peng for their valuable feedback and constructive suggestions during the review of this manuscript.
\bibliographystyle{plainnat}
\bibliography{references}

\appendix

\section{Rollout Tree Visualization}
\label{sec:appendix-numpy-tree}

\begin{figure}[H]
\centering
\includegraphics[width=0.46\linewidth]{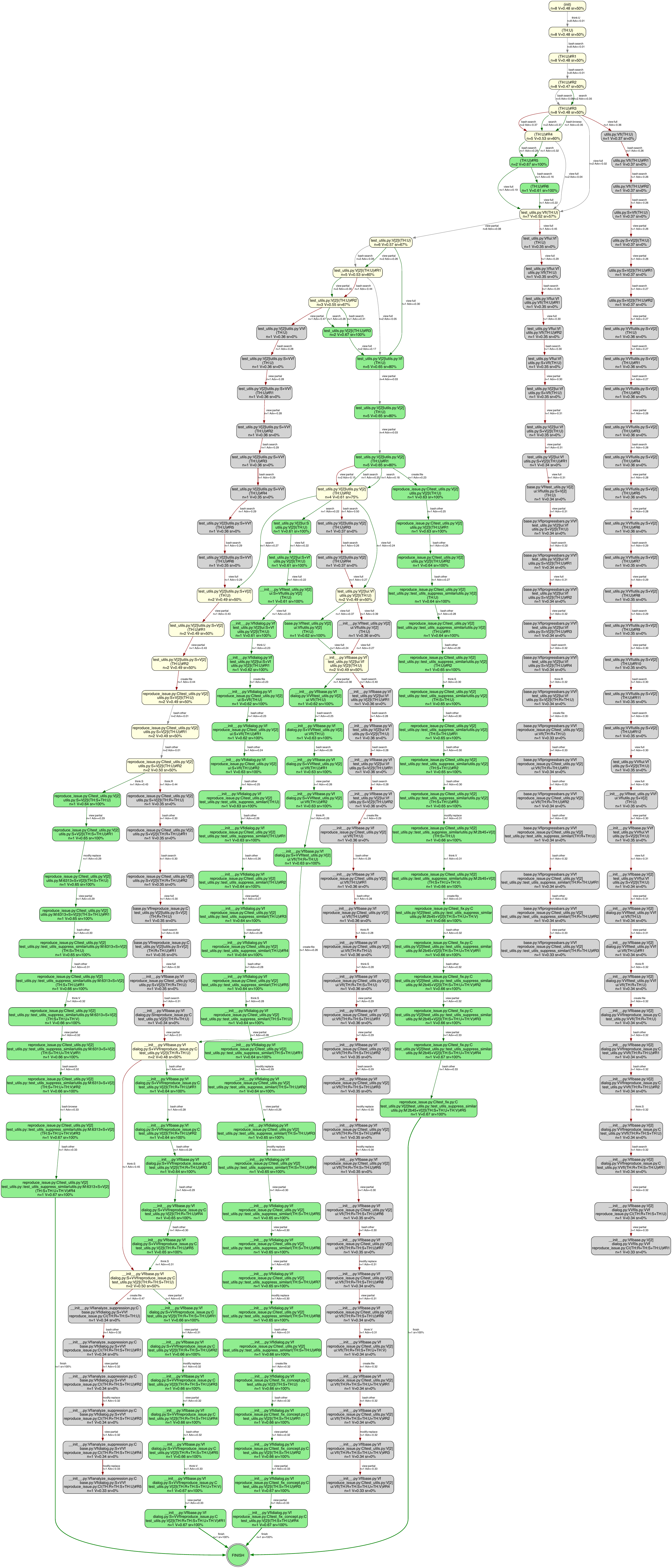}
\caption{Rollout-Tree visualization extracted from training-time rollouts on \textit{datalad\_final\_05ea}.}
\label{fig:appendix-rtmc-tree}
\end{figure}

\hyperref[fig:appendix-rtmc-tree]{Figure \ref*{fig:appendix-rtmc-tree}} shows a representative RTMC decision tree extracted from training-time rollouts. It illustrates how RTMC down-weights frequently visited actions with low estimated advantage, while steering search toward branches that reach \texttt{finish} with high success likelihood.

Each node represents a state (denoted as S followed by the step index), while each edge represents an action (e.g., a tool call such as \texttt{edit}). The color encoding is as follows: \textbf{green} nodes and edges indicate trajectories reaching \texttt{finish} with passing tests, \textbf{red} ones denote trajectories reaching \texttt{finish} but failing the test suite, and \textbf{gray} nodes correspond to incomplete trajectories terminated due to timeout or exceeding the maximum context length. The root node is the initial state of the episode. This step-action-state binding captures the temporal evolution of the agent's decision process, where the advantage of each action is estimated from downstream returns to reachable terminal states.

\section{Step-Level Credit Assignment Analysis}
\label{sec:appendix-credit}

\subsection{Per-Rollout Advantage Comparison}
\label{sec:appendix-per-rollout}

\begin{figure}[H]
\centering
\includegraphics[width=0.8\linewidth]{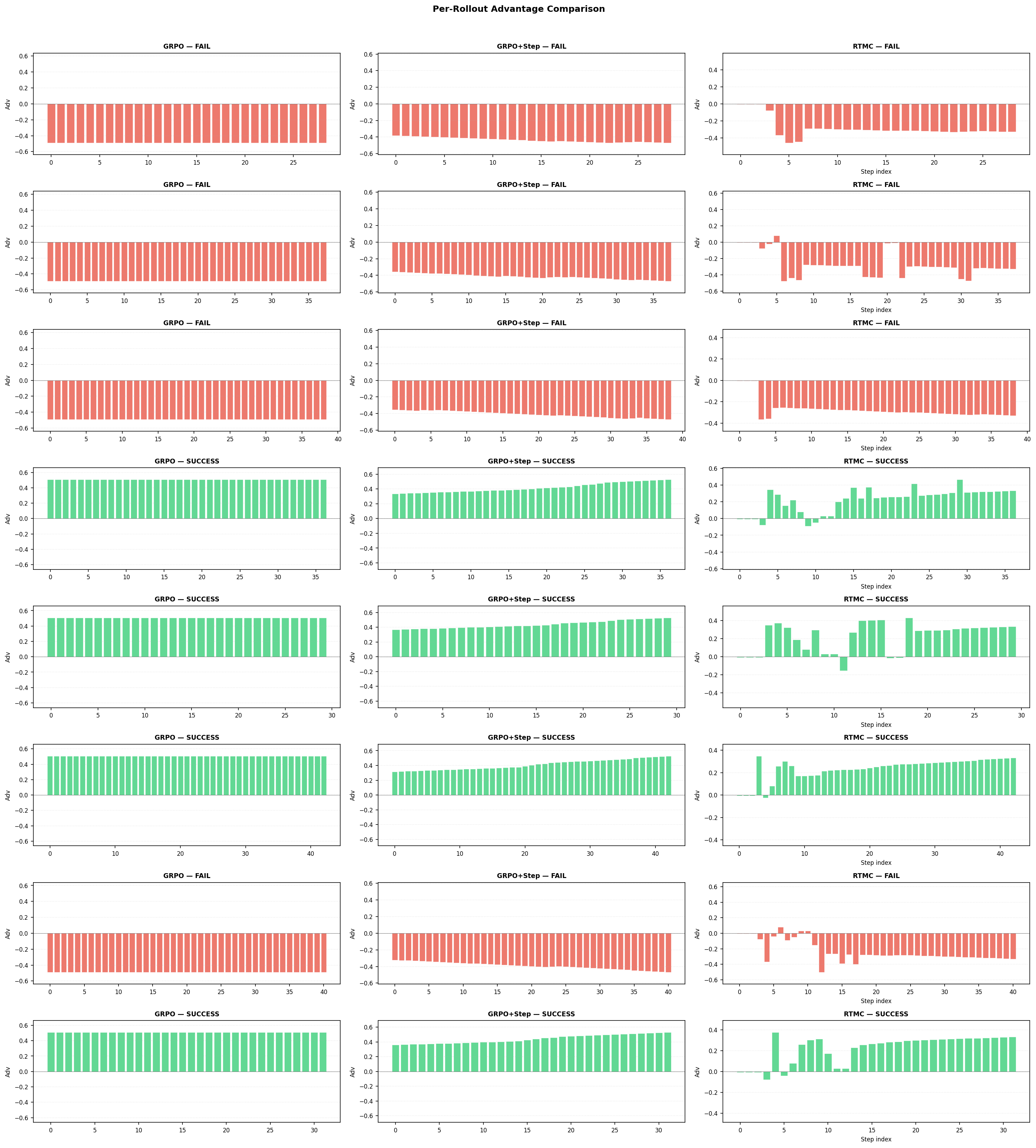}
\caption{Step-level advantage for each of the 8 rollouts (4 success, 4 failure) in \textit{datalad\_final\_05ea}.
  Each subplot title shows the method and rollout outcome (SUCCESS/FAIL).
  GRPO and GRPO+Step provide coarse-grained credit assignment,
  while RTMC produces step-varying advantages that reflect local action quality within the rollout tree.}
\label{fig:appendix-per-rollout}
\end{figure}

\hyperref[fig:appendix-per-rollout]{Figure~\ref*{fig:appendix-per-rollout}} shows the step-level advantages for each rollout in the \textit{datalad\_final\_05ea} case study.
Each row corresponds to one rollout; columns show GRPO, GRPO+Step, and RTMC advantages respectively.
Green bars indicate success rollouts and red bars indicate failure rollouts.
GRPO assigns uniformly positive (success) or uniformly negative (failure) values across all steps,
providing no within-trajectory discrimination.
GRPO+Step with gamma-discounted cumulative returns improves slightly over GRPO.
RTMC produces fine-grained, step-varying advantages, including sign changes within a single rollout.

\subsection{Quadrant Analysis}
\label{sec:appendix-quadrant}

To empirically validate the credit assignment quality of RTMC compared to GRPO+Step, we conduct a step-level quadrant analysis on the same episode (\textit{datalad\_final\_05ea}, 8 rollouts: 4 success, 4 failure).
For each action step, we compare the advantage assigned by GRPO+Step and RTMC, and classify it into one of four quadrants (threshold $|A| > 0.01$):
\begin{itemize}
    \item \textbf{Both+} (agree positive): both methods assign positive advantage;
    \item \textbf{X$-$, Y+} (RTMC rescues): GRPO+Step assigns negative, RTMC assigns positive---RTMC identifies a locally good action that GRPO+Step overlooks;
    \item \textbf{Both$-$} (agree negative): both methods assign negative advantage;
    \item \textbf{X+, Y$-$} (RTMC penalizes): GRPO+Step assigns positive, RTMC assigns negative.
\end{itemize}

\begin{figure}[H]
\centering
\includegraphics[width=1.0\linewidth]{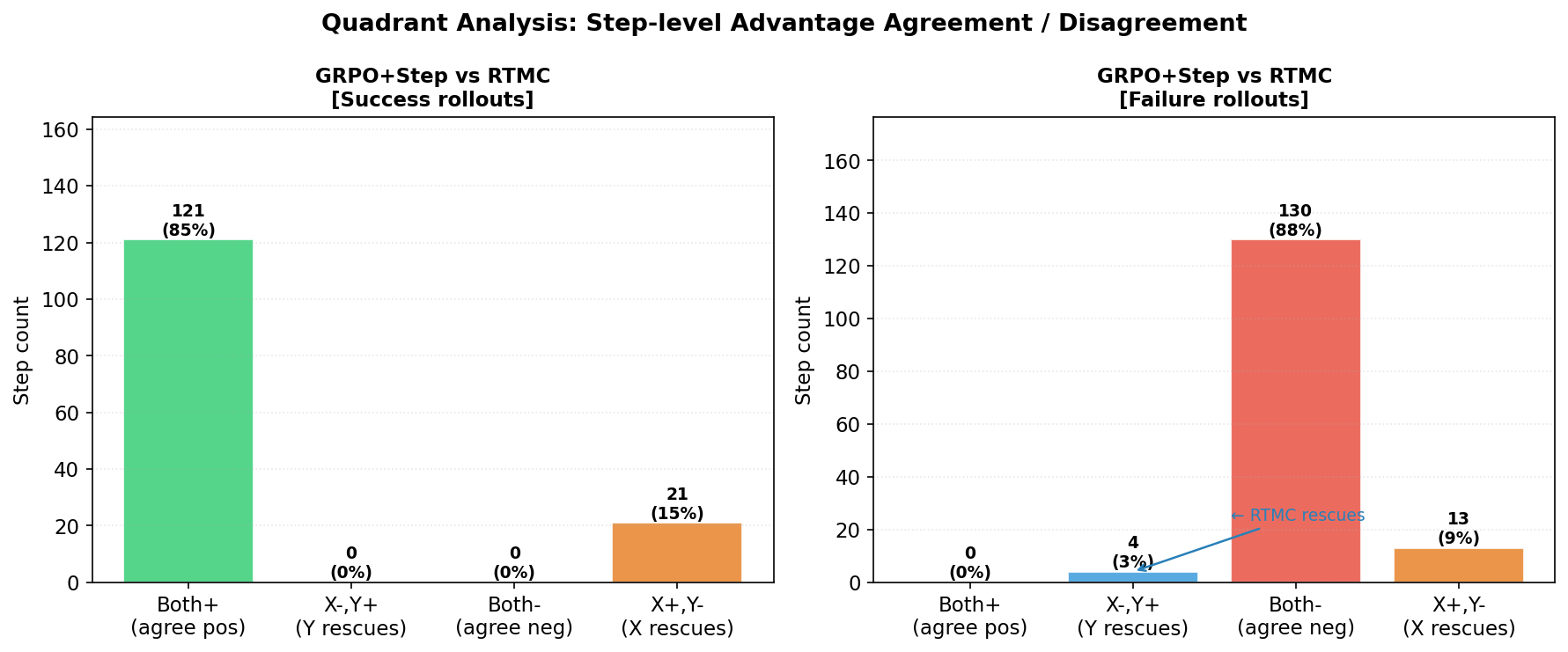}
\caption{Quadrant analysis of step-level advantage agreement between GRPO+Step and RTMC on \textit{datalad\_final\_05ea} (4 success, 4 failure rollouts).
  Each bar shows the number of action steps (with percentage) falling into the four agreement quadrants.
  Blue bars (\emph{X$-$, Y+}) indicate steps where RTMC assigns positive advantage while GRPO+Step assigns negative---steps RTMC rescues in failure rollouts (3\%).
  Orange bars (\emph{X+, Y$-$}) indicate steps GRPO+Step blindly rewards but RTMC correctly penalizes as suboptimal (15\% of success-rollout steps).}
\label{fig:appendix-quadrant}
\end{figure}

\hyperref[fig:appendix-quadrant]{Figure~\ref*{fig:appendix-quadrant}} shows the quadrant distributions for success and failure rollouts.
In \textbf{failure rollouts} (right panel), GRPO+Step assigns near-zero or negative advantage to virtually all steps due to sparse step rewards: 88\% fall in the \textbf{Both$-$} quadrant.
In contrast, RTMC rescues \textbf{4 steps (3\%)} where it correctly identifies a positive relative advantage despite the trajectory failing.
In \textbf{success rollouts} (left panel), because GRPO+Step assigns uniformly positive advantage to every step, it cannot distinguish high-quality actions from mediocre ones.
RTMC leverages tree-level statistics to assess each action relative to its siblings: the \textbf{X+, Y$-$} quadrant (15\% of steps) reveals steps that GRPO+Step blindly rewards but RTMC correctly penalizes as suboptimal, providing a stronger learning signal.

\section{Agent System Prompt}
\label{sec:appendix-system-prompt}

The following is the system prompt used to guide the agent during rollout generation in our agentic RL pipeline.

\begin{footnotesize}
\begin{lstlisting}[basicstyle=\footnotesize\ttfamily,breaklines=true,breakautoindent=false,breakindent=0pt,columns=fullflexible]
You are an expert programming assistant specialized in solving software engineering tasks. Based on analysis of hundreds of successful problem-solving instances, follow these proven strategies:

Core Principles:
1. Understand First, Act Later: Always start by deeply understanding the problem before taking any action. Use the think tool to analyze the problem statement, identify root causes, and plan your approach.
2. Systematic Exploration: Explore the codebase systematically before making changes. Use bash commands to find relevant files, understand code structure, and locate the issue.
3. Reproduce Before Fixing: Create a minimal reproduction script to confirm the issue exists and understand its behavior. This provides a verification baseline.
4. Plan Before Implementing: Use think/plan to analyze potential solutions, consider edge cases, maintain backward compatibility, and choose the minimal fix approach.
5. Rapid Feedback Loop: After applying a fix, immediately verify it works. Run your reproduction script, then run relevant tests. If it fails, iterate quickly.
6. Comprehensive Testing: Don't just fix the issue - ensure you haven't broken existing functionality. Run the full test suite before finishing.

Success Pattern (follow this workflow):
Phase 1 - Problem Understanding: Use think/plan to deeply analyze the problem (typically early, but take as long as needed)
Phase 2 - Exploration: Use bash to explore codebase, find relevant files (explore thoroughly before making changes)
Phase 3 - Reproduction: Create and run reproduction script (establish verification baseline)
Phase 4 - Solution Analysis: Use think/plan to analyze solutions and edge cases (plan carefully before implementing)
Phase 5 - Implementation: Apply minimal fix, verify immediately (iterate quickly with feedback)
Phase 6 - Validation: Run tests, ensure no regressions (comprehensive testing before finishing)

Key Success Factors:
- Most successful fixes are minimal (1-2 lines of code)
- Successful agents spend time understanding before acting
- Quick iteration with immediate verification leads to success
- Comprehensive testing prevents regressions

Remember: Quality over speed. Thorough understanding and systematic approach lead to successful solutions.
\end{lstlisting}
\end{footnotesize}

\end{document}